\documentclass[letterpaper, 10 pt, conference]{ieeeconf}  

\IEEEoverridecommandlockouts                              

\usepackage{multicol}
\usepackage[bookmarks=true]{hyperref}
\usepackage{graphicx}
\usepackage{svg}
\usepackage{amsfonts}
\usepackage{amsmath}
\usepackage{xcolor}
\usepackage{multirow}
\usepackage{booktabs}
\usepackage[ruled,linesnumbered]{algorithm2e}
\title{\LARGE \bf
Zero-Shot Constrained Motion Planning Transformers Using Learned Sampling Dictionaries
}

\author{Jacob J. Johnson$^{\dagger}$, Ahmed H. Qureshi$^{\ddagger}$, and Michael C. Yip$^{\dagger}$ 
\thanks{$^{\dagger}$J.J. Johnson and M.C.Yip are with the Electrical and Computer Engineering Department at University of California San Diego, La Jolla, CA, USA
{\tt\small \{jjj025, yip\}@eng.ucsd.edu}}
\thanks{$^{\ddagger}$ A.H. Qureshi is with the Department of Computer Science at Purdue University, West Lafayette, IN,  USA
{\tt\small ahqureshi@purdue.edu}}
}

\begin{document}
\maketitle
\thispagestyle{empty}
\pagestyle{empty}

\begin{abstract}
Constrained robot motion planning is a ubiquitous need for robots interacting with everyday environments, but it is a notoriously difficult problem to solve. Many sampled points in a sample-based planner need to be rejected as they fall outside the constraint manifold, or require significant iterative effort to correct. Given this, few solutions exist that present a constraint-satisfying trajectory for robots, in reasonable time and of low path cost. In this work, we present a transformer-based model for motion planning with task space constraints for manipulation systems. 
Vector Quantized-Motion Planning Transformer (VQ-MPT) is a recent learning-based model that reduces the search space for unconstrained planning for sampling-based motion planners.
We propose to adapt a pre-trained VQ-MPT model to reduce the search space for constraint planning without retraining or finetuning the model.
We also propose to update the neural network output to move sampling regions closer to the constraint manifold.
Our experiments show how VQ-MPT improves planning times and accuracy compared to traditional planners in simulated and real-world environments.
Unlike previous learning methods, which require task-related data, our method uses pre-trained neural network models and requires no additional data for training and finetuning the model making this a \textit{one-shot} process.
We also tested our method on a physical Franka Panda robot with real-world sensor data, demonstrating the generalizability of our algorithm. We anticipate this approach to be an accessible and broadly useful for transfering learned neural planners to various robotic-environment interaction scenarios.
\end{abstract}

\section{INTRODUCTION}
Constraint motion planning (CMP) is a fundamental challenge in robotics that involves finding a collision-free path between a start and goal configuration while satisfying certain constraints. Constraints may include kinematic constraints on the robot's joints \cite{9341283, 9551655}, dynamic constraints like torque limits \cite{safe_traj}, and task constraints like maintaining end-effector orientation \cite{10.1007/978-3-031-25555-7_17}. Trajectories that adhere to these constraints are relevant in fields such as home robotics - to move containers without spilling their content, medical robotics - constrain end-effector torques to interact with humans safely \cite{safe_traj}, and industrial robotics - articulate objects such as levers and pulleys \cite{9794598}. 

\begin{figure}[t]
    \centering
    \includegraphics[width=\linewidth]{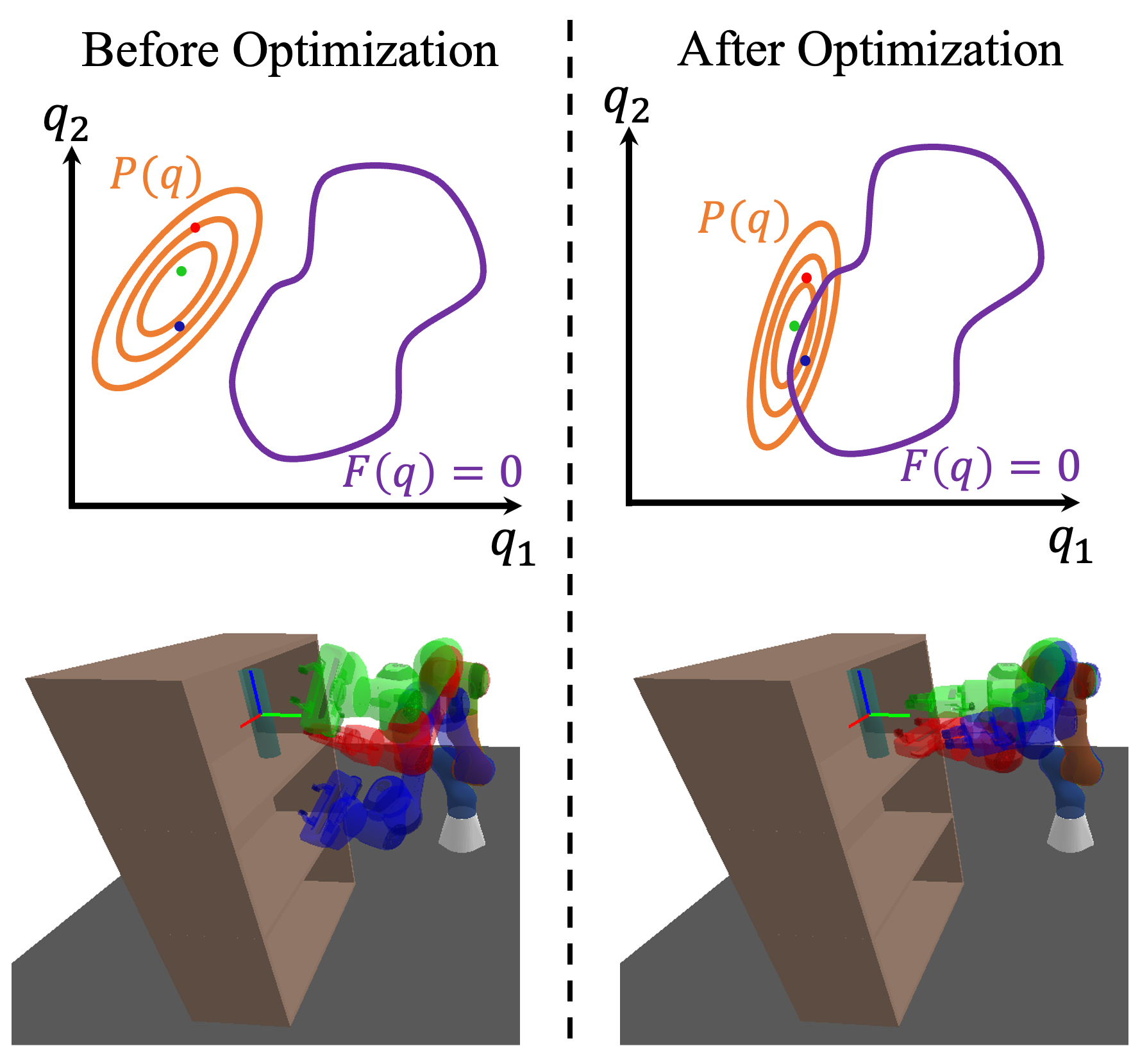}
    \vspace{-2em}
    \caption{ Concept figure showing joint configurations (bottom) sampled from distributions, $P(q)$, generated by CVQ-MPT and previous transformer-based approach \cite{johnson2023learning}. 
     CVQ-MPT uses the predicted distribution from a pre-trained VQ-MPT model (top left) and uses gradient-based optimization to update the distribution (top right) towards the constraint manifold, $F(q)=0$. Configurations sampled from this distribution are closer to the given constraint, improving planning times and sampling efficiency.}
    \label{fig:concept_fig}
    \vspace{-1.5em}
\end{figure}

The most common algorithms for finding such trajectories are sampling-based motion planners (SMP) \cite{10.5555/1213331}. These algorithms build a discrete search graph of collision-free robot states by random sampling in the planning space to connect the start and goal states. They have been particularly successful at finding solutions for robotic systems with high degrees of freedom, such as manipulators \cite{gammell2015batch}. 
Previous works have extended SMPs to plan trajectories that satisfy given constraints \cite{JAILLET2013797}. However, SMPs can be highly inefficient, especially for higher dimensional spaces. With the added complexity of satisfying constraints, the valid search region shrinks, exacerbating inefficiency.

Recent works have improved the sampling efficiency of SMPs by leveraging prior data \cite{qureshi2019motion, yu2021reducing, 8412538, 9561104} for unconstrained motion planning problems. By prudently selecting sampled points, these techniques improve planning times considerably. For CMP, the sampled points not only have to be collision-free but also need to satisfy a constraint function. Oftentimes, fully identifying these regions itself is computationally intractable. Constraint Motion Planning Networks (CoMPNet), an extension of Motion Planning Networks (MPNet), extended the idea of neural samplers for multimodal kinematic constraints \cite{9143433} to generate sampled points on this constraint manifold, i.e., regions where the constraints are satisfied. However, these methods require the collection of task-specific data to train the models.

Our previous work, Vector Quantized-Motion Planning Transformers (VQ-MPT), has shown promising results in improving the sampling efficiency of SMP planners by narrowing down the search region \cite{johnson2023learning}.  VQ-MPT has shown promising generalization capabilities in that trained models can reduce planning times for environments outside the training data. This paper introduces Constraint VQ-MPT (CVQ-MPT), a fast and efficient neural planner for kinematic and task-specific constraints. Unlike previous learning-based approaches, our model requires no additional task-related data for training or finetuning the model and uses a pre-trained VQ-MPT model for planning.
The main contributions of this paper are:
\begin{enumerate}[leftmargin=5mm]
    \item Present a zero-shot planning algorithm that requires no task-specific training data for solving constraint motion planning problems and results in 2$\times$ improvement in planning times.
    \item Formulate a gradient-based update of distributions predicted by VQ-MPT, improving planning performance.
    \item Provide empirical evidence of how trajectories generated by CVQ-MPT are 35\%-40\% shorter compared to previous SMPs, resulting in improved task execution times.
\end{enumerate}

\section{RELATED WORKS}
Numerous methods have been proposed to solve the CMP. Broadly, they can be categorized into optimization-based and sampling-based approaches. 

Optimization-based approaches formulate the entire planning problem, including constraints, as an optimization problem. In \cite{5980538}, the authors apply Covariant Hamiltonian Optimization for Motion Planning (CHOMP) \cite{5152817}, a method that uses covariant gradients to solve the unconstraint planning problem to construct unconstraint trajectories and project them onto the constraint manifold. TrajOpt \cite{10.1177/0278364914528132} improves optimization-based planning by utilizing Sequential Convex Programming (SCP) and incorporates constraint functions as penalties within the optimization problem to solve CMP. Bonalli et al. \cite{DBLP:conf/rss/BonalliCBLP19} extend SCP methods by lifting the manifold constraints to Euclidean spaces. Howell et al. \cite{8967788} use Augmented Lagrangian- Iterative Linear Quadratic Regulators (AL-iLQR) to build a rough solution and a projection-based method to refine the coarse solution to solve the constraint problem. However, these techniques require hand-tuning penalty functions for different tasks, often resulting in local minima solutions.

On the other hand, sampling-based approaches build discrete search graphs by random sampling in the planning space to connect the start and goal states. SMP planners such as Rapidly Exploring Random Trees (RRT) \cite{doi:10.1177/02783640122067453} and its variants \cite{QURESHI20151} have shown to be particularly effective in solving planning problems for robotic systems with higher degrees of freedom, such as manipulators and also easily adapt to a wide range of tasks \cite{doi:10.1146/annurev-control-060117-105226}. Sampling-based methods can solve constraint planning by generating random samples on the constraint manifold mainly through projection-based and continuation-based methods. 

The projection-based approach uses a projection operator to map sampled points onto the constraint manifold. The projection operators usually involve first-order gradients to iteratively move the sampled point toward the constraint manifold using the Jacobian of the constraint function. \cite{doi:10.1177/0278364910396389, Englert_2021} use projection-based methods for solving CMP. On the other hand, continuation-based approaches approximate constraint manifolds at local regions and use this approximation to sample points and construct local trajectories. Methods such as AtlasRRT \cite{6352929} and Tangent-Bundle RRT (TB-RRT) \cite{5980566, kim_um_suh_park_2016} simultaneously approximate the constraint manifold using tangent spaces at adhering points and uses a BiRRT to construct a tree between the start and goal points on this approximated manifold. However, the underlying optimization routines and constructing atlas's can make planning computationally expensive.



To improve the efficiency and speed of sampling-based planners, recent methods have utilized learning-based techniques to solve the planning task \cite{9718343, qureshi2019motion}. Constraint Motion Planning Networks (CoMPNet) \cite{9501956, 9143433} was the first of these methods that used a fully connected neural network to generate configurations near constraint surfaces. By selectively sampling points, these methods reduced planning times considerably. Similarly, in \cite{kicki2023fast}, a deep neural network models the constraint manifold using prior trajectory data and can compute trajectories quickly. Liu et al. \cite{pmlr-v164-liu22c} propose using reinforcement learning to find a policy that always satisfies the constraints, which allows the agent to explore the space efficiently and removes the need for a projection operator. Although these methods improve the efficiency and speed of constraint planning, they lack generalizability to new environments and require access to task-specific demonstrations. In this work, we address previous CMP drawbacks without collecting any task-specific data.

\section{Problem definition}
Consider an $n$ dimensional planning space defined by $\mathcal{C}\in\mathbb{R}^n$. 
The space is split into two subspaces $\mathcal{C}_{free}\subset\mathcal{C}$ and $\mathcal{C}-\mathcal{C}_{free} $ such that all states in $\mathcal{C}_{free}$ do not collide with any obstacle in the environment and are considered valid configurations.  
The objective of the motion planner is to generate a continuous path (or trajectory), $\sigma$, such that it connects the given start state ($q_s$) and a goal region ($\mathcal{C}_{goal}$), and all states in $\sigma$ are in $\mathcal{C}_{free}$. For CMP, all the points in $\sigma$ must also satisfy a constraint function $F:\mathcal{C}\rightarrow \mathbb{R}^k$, where $k$ is the number of constraints. For a state $q\in\mathcal{C}_{free}$ to satisfy the constraint, $F(q)=\mathbf{0}$, where $\mathbf{0}$ is a vector of zeros. This work focuses on the constraint function defined on joint configurations, $\mathcal{C}$, and not on the robot kinematics or dynamics. 

\section{BACKGROUND}
\begin{figure*}
    \centering
    \includegraphics[width=\linewidth]{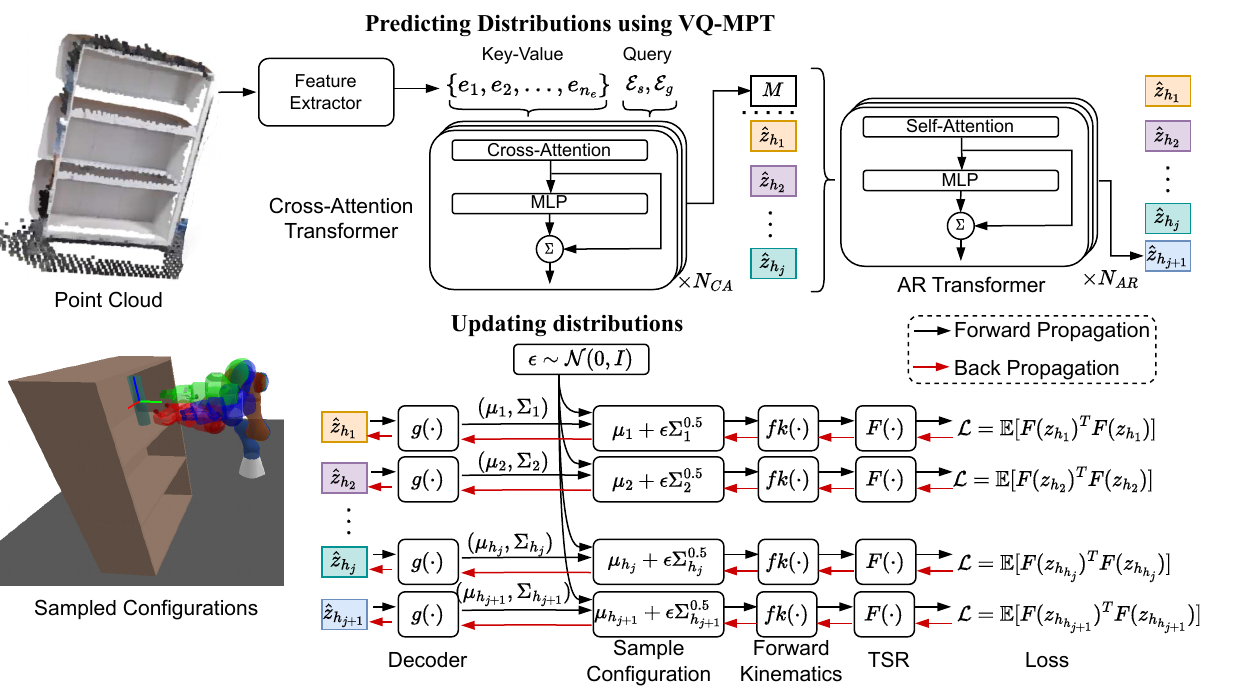}
    \vspace{-2em}
    \caption{An outline of the model architecture of CVQ-MPT. Given a point cloud, and start ($q_s$) and goal ($q_g$) configurations, a pre-trained transformer model is used to generate a set of distributions parameterized using $\{\hat{z}_{h_1}, \ldots \hat{z}_{j+1}\}$. The predicted distributions are updated using gradient-based optimization, minimizing the loss term $\mathcal{L}$, moving it closer to the constraint function, $F(q)$. This allows for sampling configurations closer to the constraint manifold.}
    \label{fig:method-idea}
    \vspace{-1em}
\end{figure*}

\subsection{Task-Space Regions}
A task space for manipulators is the space spanned by possible end-effector poses. For example, the task space for the Franka Panda arm is the $SE(3)$ space. For many applications, such as moving a cup without spilling its content, opening doors for shelves, and gasping objects from a table, constraints can be defined with respect to the task space. Task-Space Regions (TSR) \cite{doi:10.1177/0278364910396389} defines such constraints as the relative pose between the robot end-effector pose evaluated using a forward kinematics function, $fk(q):\mathcal{C}\rightarrow SE(3)$, and a given target pose. The relative pose is expressed as a vector in $\mathbb{R}^6$ where the first three elements represent the relative position ($\mathbf{p}_e^p$), and the last three represent the relative orientation ($\boldsymbol{\omega}_e^p$) in axis-angle form. Thus, the constraint function $F$ is defined using these bounds. Refer to \cite{doi:10.1177/0278364910396389} for more details.
\subsection{Vector Quantized-Motion Planning Transformer}
The VQ-MPT model consists of two stages - a quantization stage and a prediction stage. The quantization stage segments the planning space as a collection of distributions. VQ-MPT uses a Vector Quantized (VQ) model to generate the collection of distributions of the planning space. VQ models are generative models with an encoder-decoder architecture similar to Variational AutoEncoder (VAE) models but with the latent dimension represented as a collection of learnable vectors - $\mathcal{Z}_Q=\{\hat{z}_1, \hat{z}_2, \ldots, \hat{z}_N\}$ referred to as dictionaries. Each dictionary value can be reconstructed back to the planning space using a decoder function $g(\cdot)$.

The second stage generates sampling regions by predicting indexes ($\mathcal{H}$) from the dictionary set for a given planning problem and sensor data. It comprises two models - a cross-attention model with a feature extractor to embed start and goal pair, and the environment embedding into latent vectors ($M$) and a transformer-based Auto-Regressive (AR) model to predict the dictionary indexes. Transformer networks are ideal for cross-attention and AR models because they are agnostic to the environment size and can generate latent embeddings for point clouds of varying sizes and output sequences of varying lengths. The architecture of the second stage is outlined in Fig. \ref{fig:method-idea}. The environment representation (i.e., point cloud data) is passed through a feature extractor to construct the environment encodings $\mathcal{E}=\{e_1, e_2, \ldots, e_{n_e}\}$ where $e_i \in\mathbb{R}^d$ and $d$ is the size of the latent dimension. The feature extractor reduces the dimensionality of the environment representation and captures local environment structures as latent variables using set-abstraction proposed in PointNet++ \cite{NIPS2017_d8bf84be}. The start and goal states ($q_s$ and $q_g$) are projected to the start and goal embedding ($\mathcal{E}_s\in\mathbb{R}^d$ and $\mathcal{E}_g\in\mathbb{R}^d$) using a Multi-Layer Perceptron (MLP) network.
A cross-attention model uses the environment embedding, $\mathcal{E}$, and the start and goal embedding, $\{\mathcal{E}_s, \mathcal{E}_g\}$ to generate latent vectors $M$. The cross-attention model learns a feature embedding that fuses the given start and goal pair with the given planning environment. It uses the vector in $\mathcal{E}$ as key-value pairs, and $\mathcal{E}_s$ and $\mathcal{E}_g$ as query vectors to generate $M$.

The second model of stage 2 uses an AR transformer model, $\pi(\cdot)$, to predict the dictionary indexes $\mathcal{H}$. The transformer architecture allows the model to make long-horizon connections.
For each index $h_j$, the model outputs a probability distribution over $\mathcal{Z}_Q\cup \{z_g\}$ given dictionary values of previous predictions $\{\hat{z}_{h_1}, \hat{z}_{h_2}, \ldots, \hat{z}_{h_{j-1}}\}$ and the planning context $M$:
\vspace{-.5em}
\begin{equation}
\vspace{-.5em}
    \pi(h_j=i|\hat{z}_{h_1}, \ldots, \hat{z}_{h_{j-1}}, M) = p_i \quad \text{where} \ \sum_{i=1}^{N+1} p_i = 1
    \label{eqn:ar_prob}
\end{equation}
The model stops when it predicts the goal key ($z_g$).
Using the learned decoder from Stage 1, we can convert each $\hat{z}_{h_j}$ value into a Gaussian distribution in the planning space.
\begin{equation}
    g(\hat{z}_{h_j}) = \mathcal{N}(\mu_{h_j}, \Sigma_{h_j})
    \label{eqn:decoder_eqn}
\end{equation}
This work uses a pre-trained VQ-MPT model trained on unconstrained shortest trajectory data collected in simulation.

\section{CONSTRAINT VQ-MPT}
To improve the efficiency of CMP, we sample from the distributions generated by VQ-MPT and project them onto the constraint manifold. 
The following section details our sampler for constraint planning, optimization for updating predicted distributions, and planner for solving CMP.

\begin{figure*}[th]
    \centering
    \begin{minipage}{0.68\linewidth}
        \includegraphics[width=0.98\linewidth]{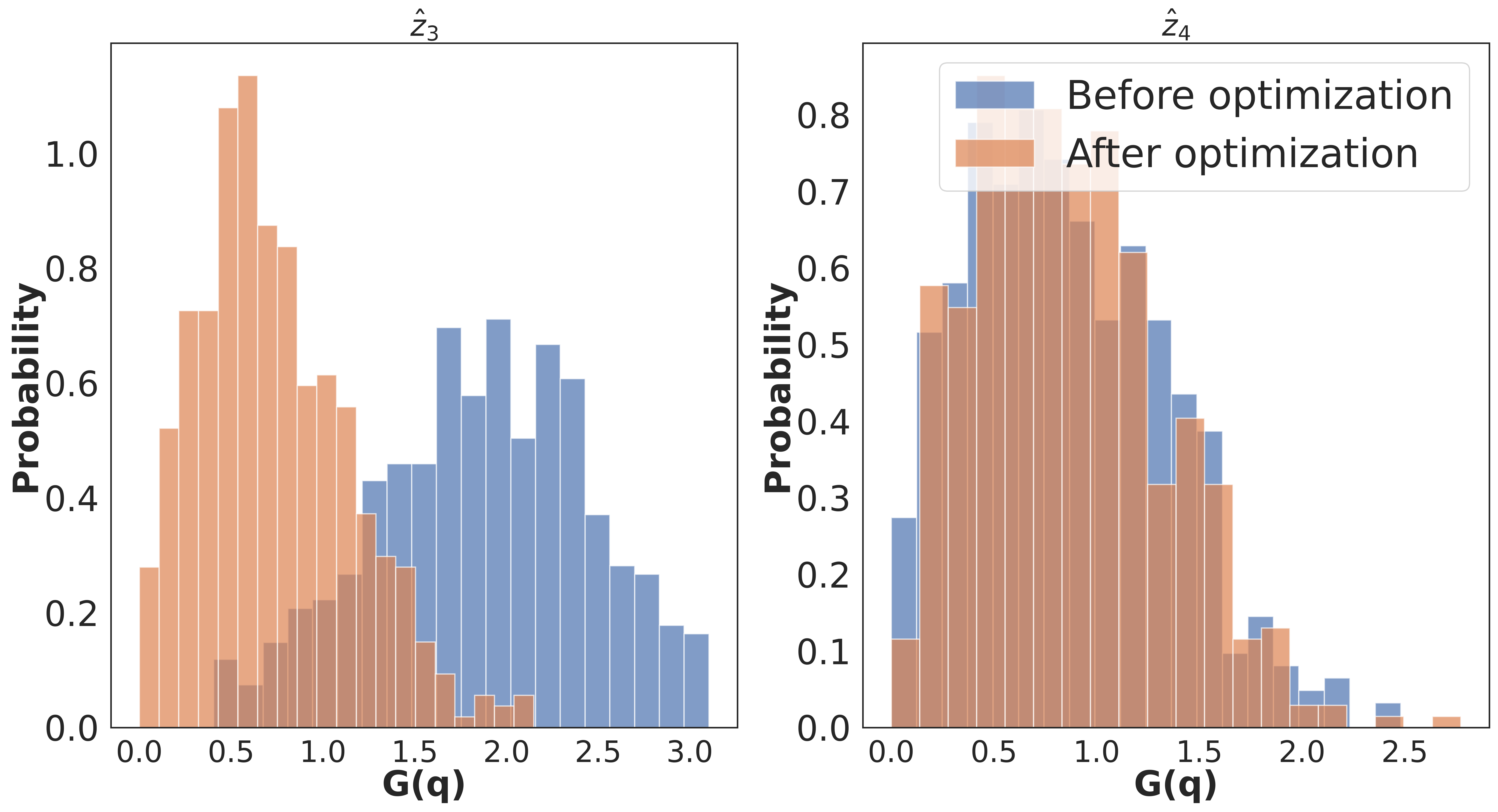}
        \vspace{-3mm}
        \caption{The histogram of the objective function, $G(q)$, before and after optimizing for two different dictionary values predicted by VQ-MPT for the place task. We can update the manifolds to generate samples closer to the constraint manifold and not push away distributions that are already closer.}
        \label{fig:optimization_dist}
    \end{minipage}
    \hfill
    \begin{minipage}{0.3\linewidth}
        \includegraphics[width=\linewidth]{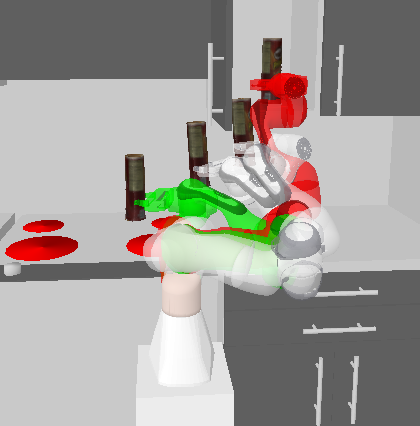}
        \caption{An example of the trajectory planned using CVQ-MPT for the place task for a given start (green) and goal (red) configurations. The constraint is to hold the can upright during the motion.}
        \label{fig:place_task}
    \end{minipage}
    \vspace{-1.5em}
\end{figure*}

\subsection{Generating samples}\label{sec:samp}
Given a start ($q_s$) and goal ($q_g$) state, we use the trained Stage 2 model of VQ-MPT to generate a sequence of dictionary indexes $\mathcal{H}=\{h_1, \ldots h_{n_h}\}$, where $h_{n_h}$ is the goal index. A beam-search algorithm similar to those used in language model tasks \cite{DBLP:conf/naacl/DevlinCLT19} is used to optimize the following:
\begin{equation}
    P(h_1, \ldots, h_{n_h} | M) = \prod_{i=1}^{n_h} \pi(h_i|h_1, \ldots, h_{i-1}, M)
    \label{eqn:prob_eqn}
\end{equation}
to generate the most probable sequence, $\mathcal{H}$. Here $\pi$ is the probability from Eqn. \ref{eqn:ar_prob}. We use a Gaussian Mixture Model (GMM) with uniform mixing coefficients reconstructed from predicted sequence $\mathcal{H}$ for sampling points. Individual Gaussian distribution is reconstructed from $\mathcal{H}$ using the decoder model from Eqn. \ref{eqn:decoder_eqn}.
\begin{equation}
    \mathcal{P}(q) = \sum_{i=1}^{n_h-1} \frac{1}{n_h - 1} g(\hat{z}_{h_i})
    \label{eqn:sample_dist}
\end{equation}
To ensure that the sampled points from $\mathcal{P}(q)$ satisfy the given constraints $F(q)$, we project them on the constraint manifold using the first-order gradient projection operator given in Algorithm \ref{alg:proj}.

\begin{algorithm}[t]
\caption{Project($q$)}\label{alg:proj}
\footnotesize
$x \gets F(q) $\;
\While{$\|x\|>\epsilon$}{
    $J\gets \nabla_q F(q)$ \;
    $\Delta q \gets -J^T(JJ^T)x$\;
    $q\gets q + \Delta q$\;
    $x \gets F(q)$\;
}
\Return{q}
\end{algorithm}

\begin{algorithm}[t]
\caption{UpdateDistribution($z$)}\label{alg:update_dist}
\footnotesize
    \Repeat{$\|\Delta z\|<\delta$}{
        $z_{cur} \gets z$ \;
        $\mu, \Sigma \gets g(z)$\;
        $\mathcal{L} \gets 0$\;
        \For{$i\leftarrow 1$ \KwTo $N$}{
            $\epsilon\sim \mathcal{N}(0, I)$ \;
            $q\gets \mu + \epsilon\Sigma^{0.5}$ \;
            $\mathcal{L} \gets \mathcal{L}+(F(q)^TF(q)$/N)\;
        }
        $\Delta z \gets \nabla_z \mathcal{L}$ \;
        $z \gets z_{cur} - \eta \Delta z$ \;
    }
\Return{z}
\end{algorithm}
\begin{algorithm}[t]
\caption{CVQMPTBiPlanner($q_{s}$, $q_{g}$, $\mathcal{P}$, $K$, $b$)}\label{alg:CVQMPTPlanner}
\footnotesize
$\tau_s \gets \{q_s\}, \tau_g \gets \{q_g\}$\;
\For{$k\gets 0$ \KwTo $K$}{
    $q_{rand}\gets$ Sample($\mathcal{P}$)\;
    $q_{near}^s\gets$ NearestNode$(q_{rand}, T_s)$ \;
    $q_{reach}^s\gets$ ConstrainedExtent$(q_{rand}, q_{near}^s, T_s)$ \;
    $q_{near}^g\gets$ NearestNode$(q_{rand}, T_g)$ \;
    $q_{reach}^g\gets$ ConstrainedExtent$(q_{rand}, q_{near}^g, T_g)$ \;
    \uIf{Connect($q_{reach}^s$, $q_{reach}^g$)}{
            $\tau \gets$ExtractPath($\tau_s, \tau_g$)\;
            \Return{Simplify($\tau$)}\;
    }
    \Else{
        Swap($\tau_s, \tau_g$)\;
    }
}
\Return{$\Phi$}
\end{algorithm}

\subsection{Improving sampling efficiency}\label{sec:opt_samp}
The samples generated from Eqn. \ref{eqn:sample_dist} are not guaranteed to lie on the constraint manifold and might require multiple optimization iterations to project them onto it. We propose a novel optimization-based update of the latent vectors in $\mathcal{H}$ such that the new distribution lies closer to the constraint, reducing projection times and consequently planning performance (Fig. \ref{fig:method-idea}). If $q^*$ adheres to the constraint surface, it must also satisfy the following:
\begin{equation}
    q^* = \min_{q} F(q)^TF(q)
\end{equation}
Thus, we can use the function $G(q)=F(q)^TF(q)$ to evaluate constraint adherence and use the same to update our latent vectors, $\hat{z}_{h_i}$. Since $G:\mathcal{C}\rightarrow \mathbb{R}^+$ is always positive, using Markov Inequality, we can upper-bound the probability of $G$ lying outside a threshold $\delta$ for the distribution $g(\hat{z}_{h_i})$.
\begin{equation}
    P(G(q)>\delta)\leq \frac{\mathbb{E}_{g(\hat{z}_{h_i})}[G(q)]}{\delta}
    \label{eqn:markov}
\end{equation}
By minimizing the upper bound in Eqn. \ref{eqn:markov}, we can implicitly reduce the samples that are further away from the constraint manifold, improving the sampling efficiency of the planner.
We can use Monte Carlo estimates of $G(\cdot)$ by sampling points on the distribution $g(\hat{z}_{h_i})$ to evaluate the upper bound in Eqn. \ref{eqn:markov}.
\begin{equation}
    \mathcal{L} = \mathbb{E}_{g(\hat{z}_h)}[G(q)] \approx \frac{1}{n}\sum_{k=1}^n G(q_k) \quad q_k\sim g(\hat{z}_h)
    \label{eqn:simplified_objective}
\end{equation}
To optimize Eqn. \ref{eqn:simplified_objective}, we want to differentiate the objective function with respect to latent variable $\hat{z}_h$. We use the reparameterization trick \cite{kingma2022autoencoding} to express the random variable $q_k$ as a function of  $\mu_{h}$, $\Sigma_{h}$, and a normal distribution $\mathbf{\epsilon}$.
\begin{equation}
    q_k = \mu_{h} + LD \mathbf{\epsilon}\qquad \mathbf{\epsilon}\sim \mathcal{N}(0, I),\ \Sigma_{h} = LD^2L^T
    \label{eqn:reparam}
\end{equation}
where $L$ and $D$ are a lower triangular and diagonal matrix, respectively. To generate points on $g(\hat{z}_h)$, we can sample points on $\mathcal{N}(0, I)$ and transform it using Eqn. \ref{eqn:reparam}. By substituting Eqn. \ref{eqn:reparam} in Eqn. \ref{eqn:simplified_objective}, the gradient of $\mathcal{L}$ becomes a function of deterministic parameters.

\begin{equation}
    \nabla_{\hat{z}_h} \mathcal{L} \approx \frac{1}{n}\sum_{k=1}^n \nabla_{q} F(q_k)\frac{\delta (\mu_{h_j} + LD \mathbf{\epsilon}_k)}{\delta \hat{z}_h}
    \label{eqn:obj_derivate}
\end{equation}
Using the derivates from Eqn. \ref{eqn:obj_derivate}, any optimization methods, such as Gradient Descent (GD), can be used to minimize the upper bound. Algorithm \ref{alg:update_dist} provides a general outline for updating the latent variable. Fig. \ref{fig:optimization_dist} shows the histogram of the objective function before and after optimization.

\subsection{Planning}
Any SMP can generate the trajectory using the sampling strategy from Section \ref{sec:samp} and \ref{sec:opt_samp}. We provide Algorithm \ref{alg:CVQMPTPlanner}, a bidirectional planning algorithm, to generate a path using samples from the distribution in Eqn. \ref{eqn:sample_dist}. The \textit{CVQMPTBiPlanner} function takes the start and goal state ($q_s$ and $q_g$), the GMM model ($\mathcal{P}$), the number of samples to generate ($K$), and a threshold value ($b$) to sample the goal state and returns a valid trajectory. The function \textit{NearestNode} finds the closest node on the tree to the sampled node, while the function \textit{ConstrainedExtend} extends the tree from the nearest node toward the sampled configuration until a collision or constraint violation occurs. For start and goal regions defined by TSR, which occur during grasping or object-placement tasks, we sample a pose from a TSR and use an Inverse Kinematics (IK) solver \cite{7363472} to generate a valid collision-free configuration. This configuration is used to identify the sampling region using the VQ-MPT planner. Since the VQ-MPT planner can generate paths rapidly, multiple configurations can be evaluated successively to find a valid path.

In \cite{doi:10.1177/0278364910396389, doi:10.1177/0278364919868530}, the authors have proved the probabilistic completeness of projection-based planners. They show that any RRT-based algorithm with a non-zero probability of sampling in an $n$-dimensional ball centered at a node of the existing tree together with the projection operator in Algorithm \ref{alg:proj} is probabilistically complete. Since we use a set of Gaussian Distributions to sample, which spans the entire planning space, our planner also satisfies these conditions, making it probabilistically complete as well.

\section{EXPERIMENTS}

\begin{figure*}[t!]
    \centering
    \includegraphics[width=\linewidth]{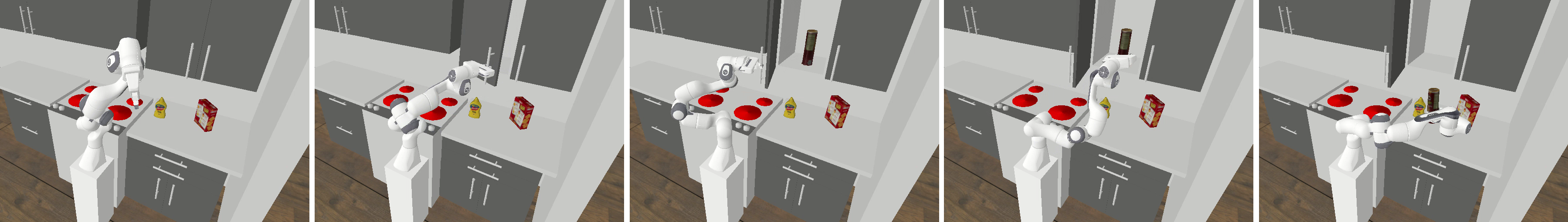}
        \vspace{-2em}
    \caption{Snapshots of the trajectory planned using CVQ-MPT for the Panda Arm completing the multi-sequence task (from left to right). CVQ-MPT can plan trajectories for various types of planning constraints in complex environments.}
    \label{fig:multi_seq}
    \vspace{-0.5em}
\end{figure*}

\begin{figure}[t!]
    \centering
    \includegraphics[width=\linewidth, trim={0 0.5in 1in 0in},clip]{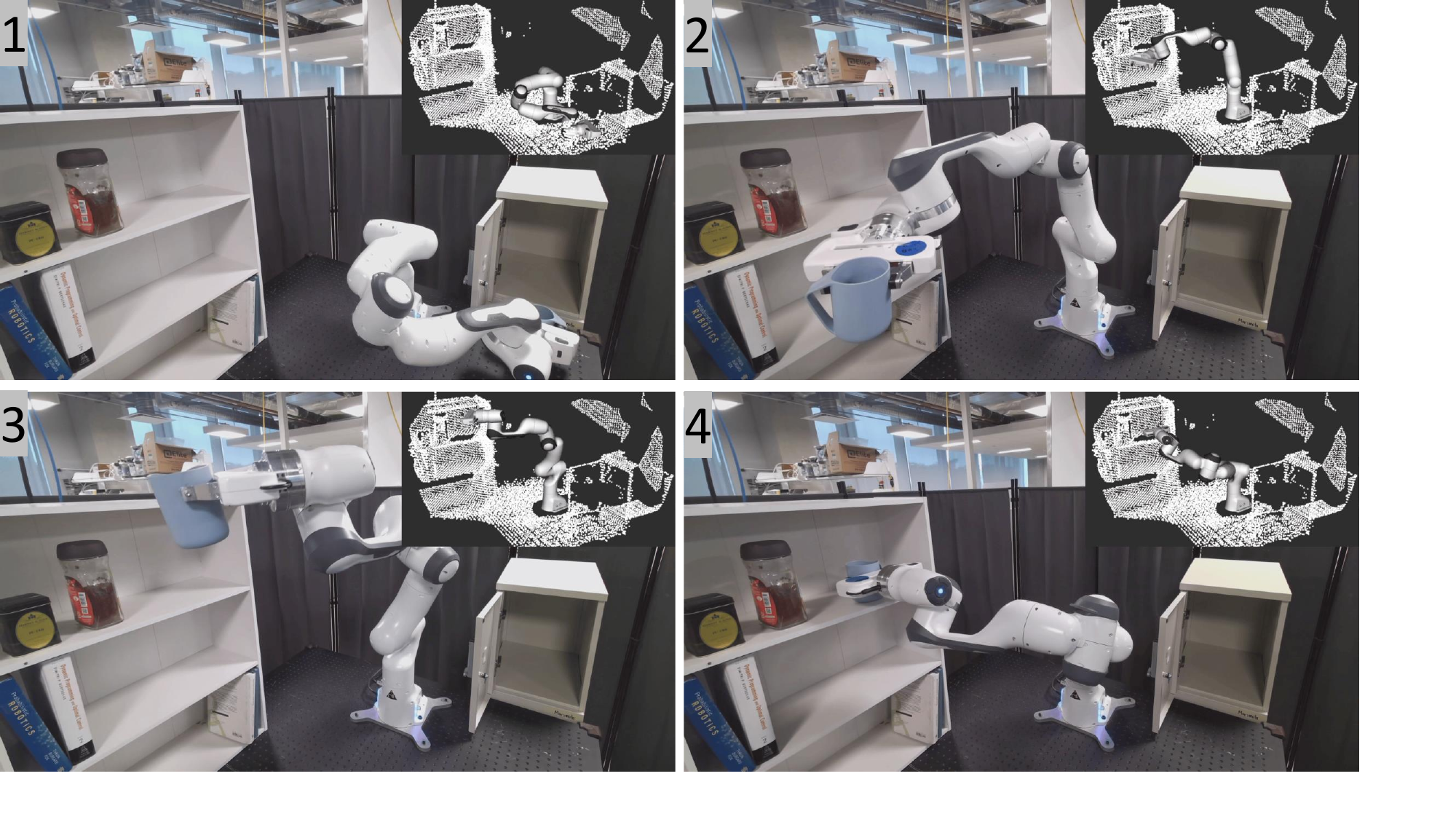}
        \vspace{-2em}
    \caption{Sequences of the trajectory planned using CVQ-MPT for the Panda Arm on the physical robot and the corresponding point cloud used to represent the environment (1$\rightarrow$2$\rightarrow$3$\rightarrow$4). CVQ-MPT can plan trajectories using physical sensor data and achieve the same level of performance observed in simulated environments.}
    \label{fig:realworld_task}
\end{figure}


\begin{table*}[t]
    \caption{Place Task and Physical Robot Experiments: accuracy, planning time, vertices, \& path length}
    \vspace{-3mm}
    \centering
    \scriptsize
    \begin{tabular}{ccccccccc}
        \toprule
        Environments & & CBiRRT & Atlas-RRT & TB-RRT & MPNet (w/ proj) & CVQ-MPT & Opt-CVQ-MPT \\
        \midrule
        \multirow{4}{*}{Place Task} & Accuracy    & 93.7\% & 100\% & 89.76\% & 33.85\% & 97.63\% & 98.42\% \\
                                    & Time (s)    & 22.95  & 21.80 & 25.31   & 20.61   & 11.25   & 11.03   \\
                                    & Vertices    & 40     & 37    & 32      & 35.98   & 26      & 24      \\
                                    & Path Length & 8.715  & 5.85  & 7.28    & 8.934   & 5.080   & 5.507   \\
        \midrule
        \multirow{4}{*}{Physical robot} & Accuracy    & 10/10  & 10/10   & 10/10   & 4/10   & 10/10   & 10/10   \\
                                        & Time (s)    & 8.37   & 11.01   & 13.98   & 7.35   & 9.89    & 8.37   \\
                                        & Vertices    & 38     & 28      & 34      & 5      & 56      & 41      \\
                                        & Path Length & 12.643 & 10.613  & 12.659  & 18.747 & 7.530   & 7.313   \\
        \midrule
        \bottomrule
        \end{tabular}
        \label{tab:stats_table}
    \vspace{-1.25em}
\end{table*}
\setlength\tabcolsep{4pt}
\begin{table}[t]
    \caption{Multi-sequence task: planning, executing times}
    \centering
    \vspace{-2mm}
    \scriptsize
    \begin{tabular}{ccccccc}
        \toprule
        & CBiRRT & Atlas-RRT & TB-RRT & CVQ-MPT & Opt-CVQ-MPT \\
        \midrule
        Planning (s)     & 19.71  & 23.46 & 19.24 & 13.10   & 10.98   \\
        Execution (s)  & 22.11  & 20.48 & 22.49 & 20.86   & 18.66   \\
        \midrule
        Total Time (s)          & 41.82  & 43.94 & 41.73 & 33.96   & 29.64    \\
        \midrule
        \bottomrule
        \end{tabular}
        \label{tab:task_stats}
    \vspace{-2em}
\end{table}

This section covers our experiment setup and compares CVQ-MPT with three other planners - CBiRRT \cite{doi:10.1177/0278364910396389}, Atlas-RRT \cite{6352929}, and TB-RRT \cite{5980566} for a 7D Franka Panda arm on simulated and physical scenes. We also compared the performance of a pre-trained MPNet \cite{qureshi2019motion}, which used the projection operator for steering.

\subsection{Setup}
We used pre-trained VQ-MPT and MPNet models from our prior work \cite{johnson2023learning} for a 7D Franka Panda robot arm. The model was trained on simulated data using unconstrained trajectories. No constraint trajectories were used to finetune these models. All planners used in this work were implemented using the Open Motion Planning Library (OMPL) \cite{sucan2012the-open-motion-planning-library} on a system with an AMD Ryzen Threadripper 1950X CPU with an Nvidia RTX 3090 GPU.

\subsection{Place Task}

First, we compared our planner's performance against traditional and learning-based SMP algorithms in solving a placement task (Fig \ref{fig:place_task}). A can is randomly placed on the kitchen counter, and the objective is to plan a path to place it on the shelf without tilting it. To quantify planning performance, we measured four metrics: planning time - the time it takes for the planner to generate a valid trajectory; vertices - the number of collision-free vertices required to find the trajectory, accuracy - the percentage of planning problems solved before a given cutoff time, and path length - the sum of all the Euclidean distance between adjacent joint states. The number of vertices will help us to determine the efficiency of different constraint planning techniques since projecting points on the constraint manifold can be computationally expensive \cite{doi:10.1146/annurev-control-060117-105226}. We tested 120 different planning problems; Table \ref{tab:stats_table} summarizes the results.

CVQ-MPT and Opt-CVQ-MPT achieve similar or better accuracy than previous constraint planners while planning shorter trajectories. CVQ-MPT produces shorter paths because the underlying VQ-MPT model was trained to identify sampling regions in $\mathcal{C}$ where the shortest unconstrained path exists. Thus, our planner projects the unconstrained shortest path on the constraint manifold. We can also observe that by reducing the search space, CVQ-MPT can plan constraint trajectories almost 2$\times$ faster than previous planners. The MPNet model on the other hand achieves poor accuracy since the model is not generalizable to unseen environments \cite{johnson2023learning}.

\subsection{Multi-Sequence Task}
Our next experiment compared the planning and execution of CVQ-MPT for a sequence of tasks with varying constraints. The task involves opening the kitchen cabinet, grasping a can from the shelf, and placing it on the kitchen counter (Fig. \ref{fig:multi_seq}). We compared 10 different tasks, where the robot's start position and the can's final goal position were random. Due to the poor performance of MPNet on the place task, we did not compare against it for this experiment. The average total planning and executing time results for the tasks are reported in Table \ref{tab:task_stats}. CVQ-MPT reduces total planning time by 45\%, while planning a shorter path can reduce task execution times by 18\%. This also shows how our planner can adapt to diverse types of task constraints.

\subsection{Real-world environment}
To evaluate the performance of our planner on physical sensor data, we tested it in a real-world environment (Fig. \ref{fig:realworld_task}). The environment was represented using point cloud data from Azure Kinect sensors, and collision checking was done using the Octomap collision checker from Moveit \cite{Coleman2014ReducingTB}. 
Camera to robot base transform was estimated using marker-less pose estimation technique \cite{lu2023markerless}. 
We tested on 10 random start and goal configurations, and the results are summarized in Table \ref{tab:stats_table}. We observe that CVQ-MPT and Opt-CV-MPT outperform traditional planners, similar to the Place Task experiment. 
This experiment shows that CVQ-MPT models can also generalize well to physical sensor data without further training or fine-tuning. Such generalization will benefit the larger robotics community since other researchers can use trained models in diverse settings without collecting new data or fine-tuning the model.


\section{CONCLUSION}
We introduced CVQ-MPT, a zero-shot planning framework for solving constraint motion planning problems in this work. By reducing the search space for sampling-based planners, we improve constraint planning efficiency and speed, simultaneously generating shorter trajectories than previous planners. We further refine our search space by optimizing the predicted distributions to be closer to the constraint manifold, further improving planning performance. 
As we have shown, CVQ-MPT also improves task execution times and success rate which will enable future robotic systems to handle more complex tasks with intricate planning sequences. Future works can explore the application of CVQ-MPT to such task and motion planning problems.

\bibliographystyle{IEEEtran}
\bibliography{references}

\end{document}